\documentclass[conference]{IEEEtran}

\usepackage{graphicx}
\usepackage{booktabs}
\usepackage{amsmath}
\usepackage{cite}
\usepackage{url}
\usepackage{hyperref}

\hypersetup{
  colorlinks=true,
  linkcolor=black,
  citecolor=black,
  urlcolor=blue
}

\begin{document}

\title{BHDD: A Burmese Handwritten Digit Dataset}

\author{
  \IEEEauthorblockN{Swan Htet Aung\textsuperscript{\dag}, Hein Htet, Htoo Say Wah Khaing, Thuya Myo Nyunt}
  \IEEEauthorblockA{Expa.AI Research Team\\Yangon, Myanmar\\
  \textsuperscript{\dag}Corresponding author: swan@expa.ai}
  \thanks{This dataset was collected and analyzed in 2019. We release this
  preprint to establish a citable reference for the community.}
}

\maketitle

% ------------------------------------------------------------------
\begin{abstract}
We introduce the Burmese Handwritten Digit Dataset (BHDD), a collection
of 87,561 grayscale images of handwritten Burmese digits in ten classes.
Each image is $28\times28$ pixels, following the MNIST format. The
training set has 60,000 samples split evenly across classes; the test
set has 27,561 samples with class frequencies as they arose during
collection. Over 150 people
of different ages and backgrounds contributed samples. We analyze the
dataset's class distribution, pixel statistics, and morphological
variation, and identify digit pairs that are easily confused due to the
round shapes of the Myanmar script. Simple baselines (an MLP, a two-layer CNN, and an improved
CNN with batch normalization and augmentation) reach 99.40\%,
99.75\%, and 99.83\% test accuracy respectively.
BHDD is available under CC BY-SA 4.0 at
\url{https://github.com/baseresearch/BHDD}.
\end{abstract}

\begin{IEEEkeywords}
handwritten digit recognition, benchmark dataset, Myanmar script,
Burmese digits, optical character recognition
\end{IEEEkeywords}

% ------------------------------------------------------------------
\section{Introduction}

MNIST~\cite{lecun1998gradient} set the standard for handwritten digit
recognition in 1998: 70,000 images of Latin digits, each $28\times28$
grayscale pixels, with a fixed train/test split. It has been used in
thousands of studies since. EMNIST~\cite{cohen2017emnist} extended the
same format to Latin letters, and
Kuzushiji-MNIST~\cite{clanuwat2018deep} showed that Japanese cursive
characters produce very different error patterns than Latin
digits; different scripts need their own benchmarks.

Myanmar has roughly 55~million people. Burmese, the main language, is
spoken natively by over 33~million. The Myanmar script, called
\textit{sar-lone} (literally ``round script''), developed its circular
letterforms because it was traditionally written on palm
leaves, where straight strokes would tear through the leaf. The digits carry
over this roundness: they are built from curves, loops, and arcs, and
several pairs look similar enough to confuse classifiers.

Until now, no public benchmark dataset existed for Burmese handwritten
digits. We release BHDD to fill that gap. The dataset contains 87,561
labeled images from over 150 contributors. We also provide a detailed
analysis of the data and three baseline models.

% ------------------------------------------------------------------
\section{Related Work}

MNIST~\cite{lecun1998gradient} has 70,000 handwritten Latin digit images
and remains the most common benchmark for digit recognition. Its fixed
format ($28\times28$ grayscale, integer labels, standard split) made it
easy to compare results across models.
EMNIST~\cite{cohen2017emnist} added handwritten letters in the same
format.

Kuzushiji-MNIST~\cite{clanuwat2018deep} took the same approach for
cursive Japanese characters and found that the error patterns looked
nothing like those on Latin digits---different scripts need their own
benchmarks. Since then, handwritten digit datasets have appeared for
Kannada~\cite{prabhu2019kannada}, Bengali~\cite{alam2018numtadb},
Arabic~\cite{elsawy2017arabic}, and
Persian~\cite{khosravi2007hoda}.

Myanmar script recognition has been studied since at least 2005, when
Sandar~\cite{sandar2005comparison} compared off-line handwriting and
print recognition using hidden Markov models.
Lwin and Wu~\cite{lwin2020myanmar} later used K-means clustering with a
CNN to handle visual similarity among Myanmar's glyphs, and
Aung et al.~\cite{aung2024myocr} released myOCR, an OCR pipeline
for machine-printed Myanmar text. However, none of these efforts
produced a publicly available benchmark for handwritten digits.
BHDD is the first.

% ------------------------------------------------------------------
\section{Dataset}

% ---- 3.1 ----
\subsection{Collection Methodology}

\begin{figure}[!t]
  \centering
  \includegraphics[width=\columnwidth]{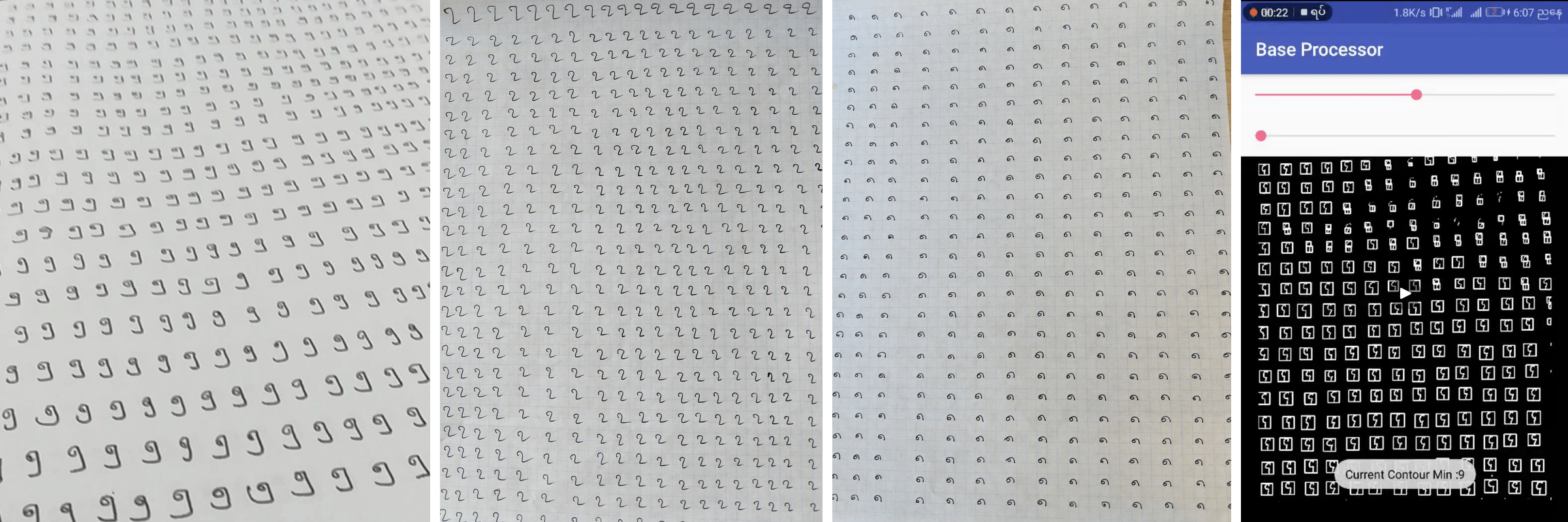}
  \caption{Sample collection sheets on plain paper (left) and grid
  paper (middle two), and the mobile preprocessing app (right) that
  let contributors verify digit extraction before submission.}
  \label{fig:sheets}
\end{figure}

We organized a community collection effort through the Expa.AI Research
Team. Over 150 people wrote Burmese digits on plain A4 paper (both white
and yellow sheets), each filling ten or more pages (one per digit
class), with roughly 500--600 digits per page. In total, approximately
2,500 sheets were collected. Contributors came from Taungoo Computer University
(volunteers and interns), St.\ Augustine / B.E.H.S~(2) Kamayut (high
school students), and friends and family of the research team. Ages
ranged from teenagers to people in their 50s, with occupations including
clerks, programmers, and students. Fig.~\ref{fig:sheets} shows
representative collection sheets.

The majority of sheets were photographed with phone cameras rather
than scanned. Variable lighting, angles, and camera quality made
automated digit extraction difficult, so we developed an Android
application (Fig.~\ref{fig:sheets}, right) that applied adaptive
thresholding and contour detection in real time, capturing per-photo
preprocessing parameters as metadata. Contributors could adjust
settings and retake photos until the extracted digits were clean.
Because extraction yield from phone photos was low, the team collected
substantially more sheets than the final dataset required.

The extraction pipeline used OpenCV: convert to grayscale, binarize
with adaptive thresholding (using the per-photo parameters captured by
the app), find contours to locate each digit, crop to the bounding box,
center in a $28\times28$ frame, and normalize pixel intensity.

Quality assurance was a two-stage process: the 20-member annotation
team inspected samples distributed via a chatbot-based review tool,
and two data engineers then performed a final verification pass over
the complete dataset. Mislabeled, illegible, and duplicate images were
removed at each stage. All 87,561 images in the final dataset are
verified unique: no exact duplicates exist within or across the
training and test sets.

Approximately 120 contributors were assigned to the training set and
the remaining roughly 30 to the test set, so no writer's handwriting
appears in both splits. Most contributors were based in Yangon, with
smaller groups from Mandalay, Nay Pyi Taw, Shan State, and the
United States.

% ---- 3.2 ----
\subsection{Data Format and Structure}

\begin{figure}[!t]
  \centering
  \includegraphics[width=\columnwidth]{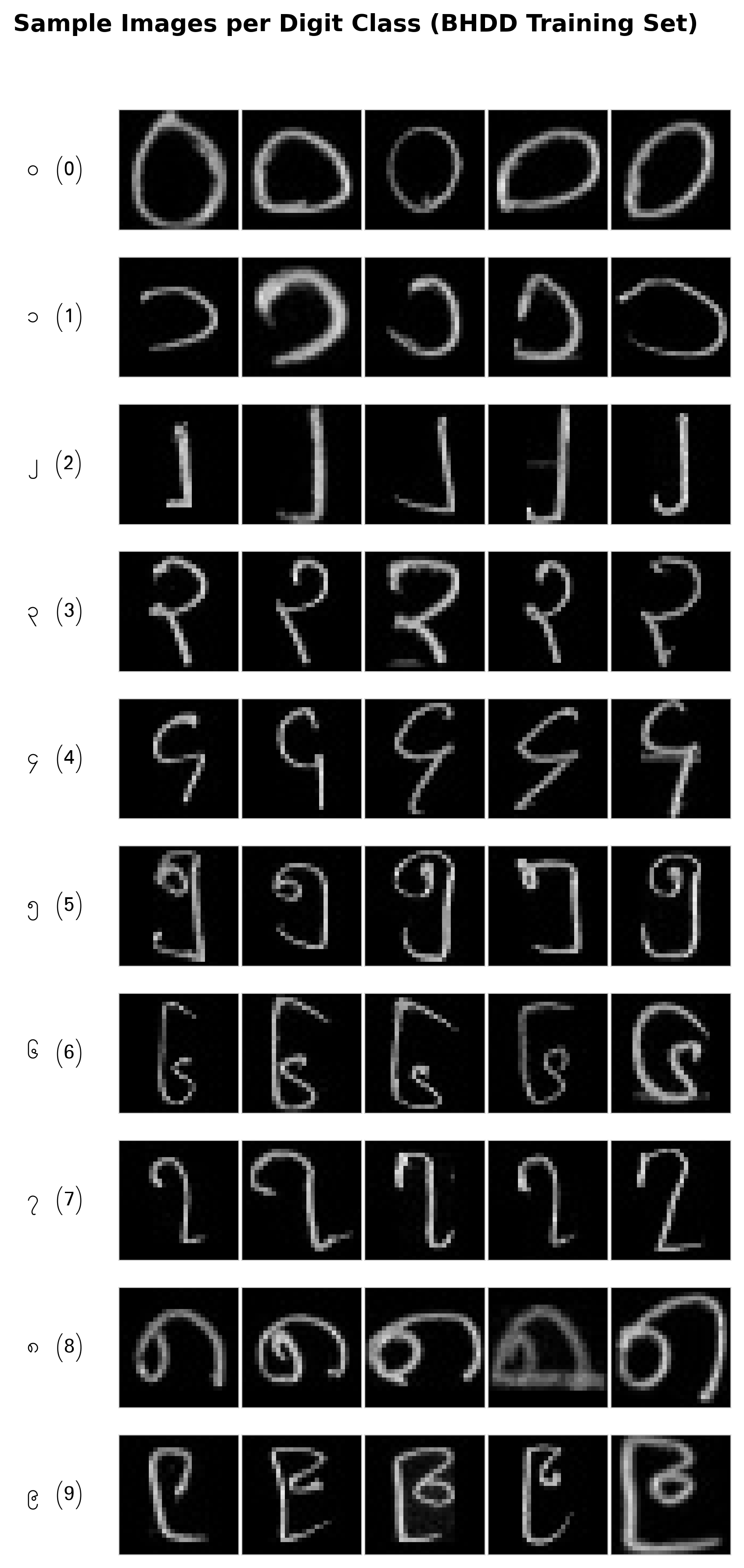}
  \caption{Samples from all ten Burmese digit classes (rows: 0--9).}
  \label{fig:sample_grid}
\end{figure}

Each sample is a $28\times28$ grayscale image (uint8, values 0--255)
with an integer label from 0 to 9. The dataset is available in Python
pickle and gzip-compressed IDX formats; the IDX files work directly with
any MNIST data loader. Fig.~\ref{fig:sample_grid} shows samples from all
ten classes.

% ---- 3.3 ----
\subsection{Class Distribution}

\begin{figure}[!t]
  \centering
  \includegraphics[width=\columnwidth]{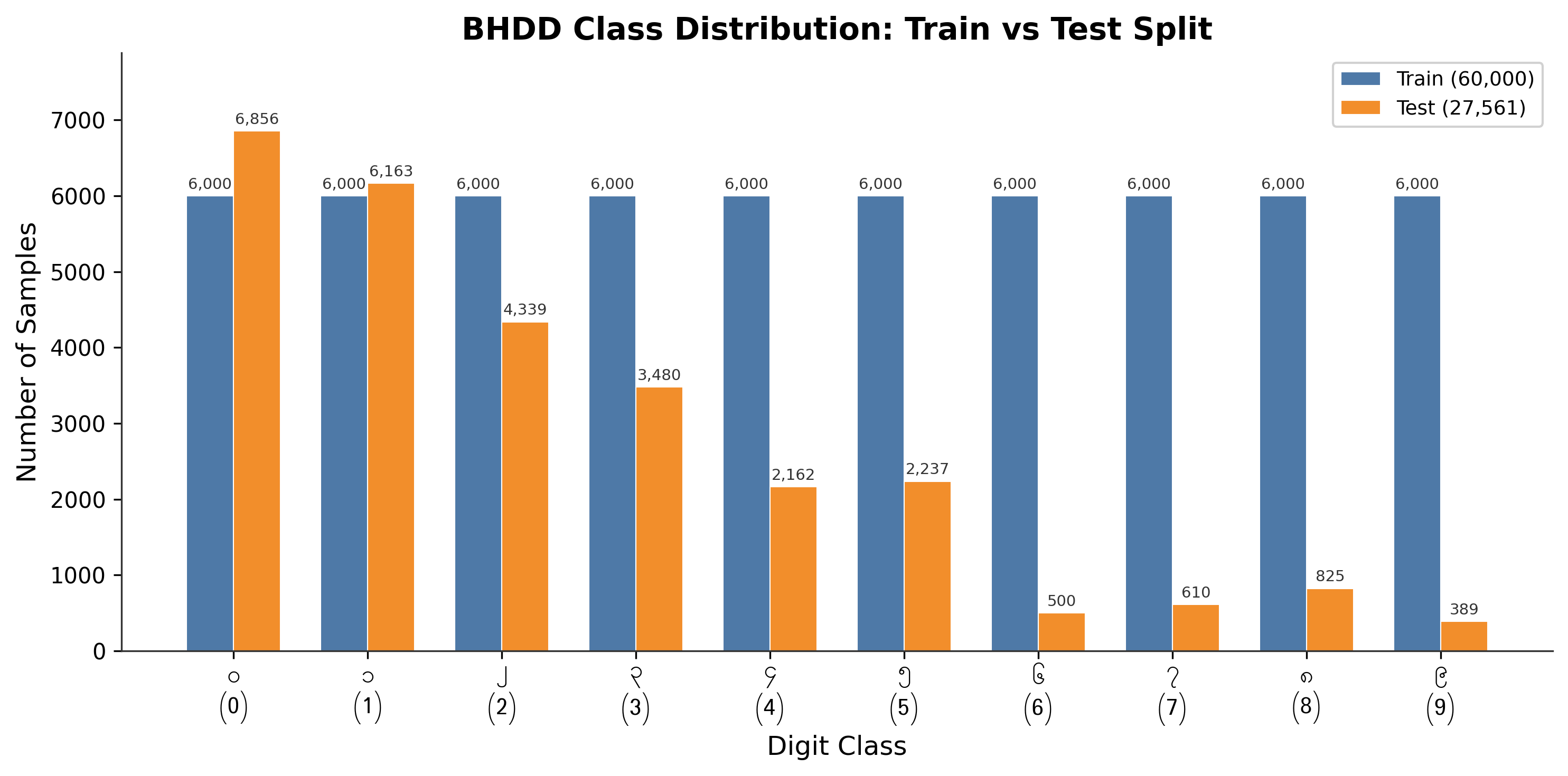}
  \caption{Training set (balanced, 6,000 per class) vs.\ test set
  (unbalanced, 6,856 down to 389).}
  \label{fig:class_dist}
\end{figure}

The dataset contains 87,561 images: 60,000 for training and 27,561
for testing. The split is by contributor; all pages from a given person
go entirely to one set, so no writer's handwriting appears in both.
The training set is downsampled to exactly 6,000 samples per class.
The test set is left unbalanced (Fig.~\ref{fig:class_dist}), with
per-class counts ranging from 6,856 (class~0) to 389 (class~9),
reflecting how often each digit appeared in the collected sheets.

% ---- 3.4 ----
\subsection{Visual and Statistical Analysis}

\begin{figure}[!t]
  \centering
  \includegraphics[width=\columnwidth]{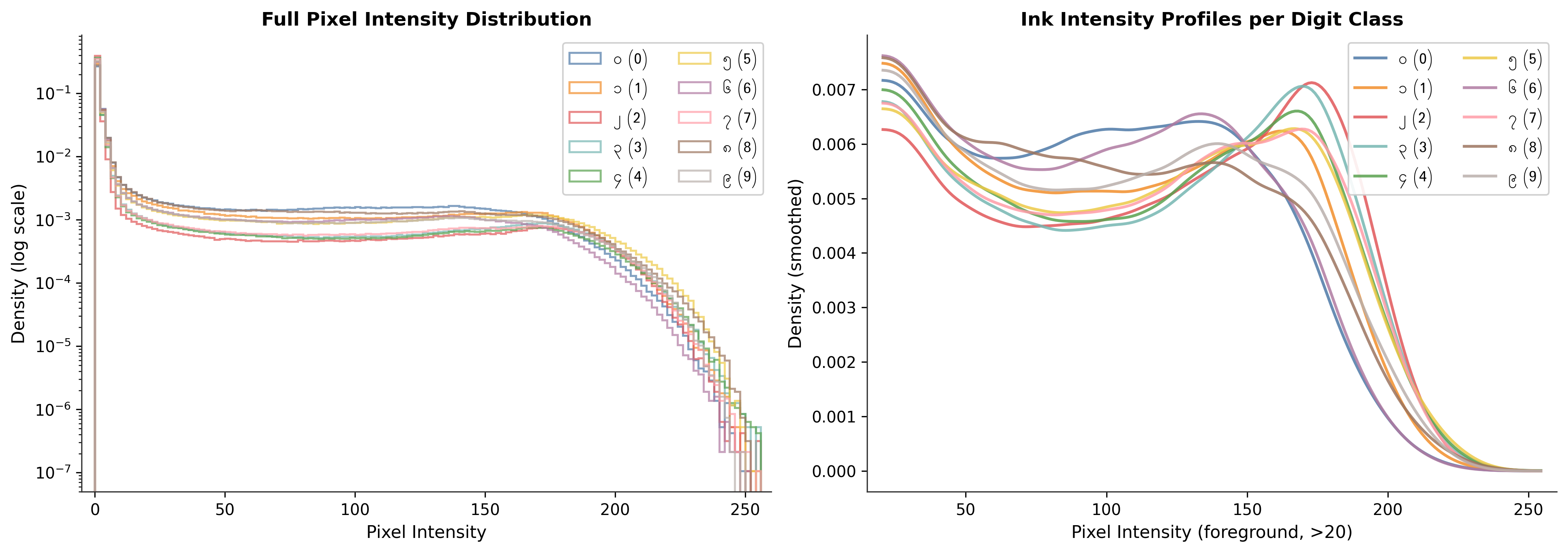}
  \caption{Pixel intensity distributions per class. Left: log-scale
  histogram. Right: per-class kernel density estimates.}
  \label{fig:pixel_intensity}
\end{figure}

\textbf{Pixel intensity.} Mean pixel intensity per class ranges from
12.5 (class~2) to 26.5 (class~0). Ink coverage (the fraction of
non-zero pixels) ranges from 30.4\% (class~2, a thin hook) to 56.8\%
(class~0, a full circle). Per-class distributions are shown in
Fig.~\ref{fig:pixel_intensity}.

\begin{figure}[!t]
  \centering
  \includegraphics[width=\columnwidth]{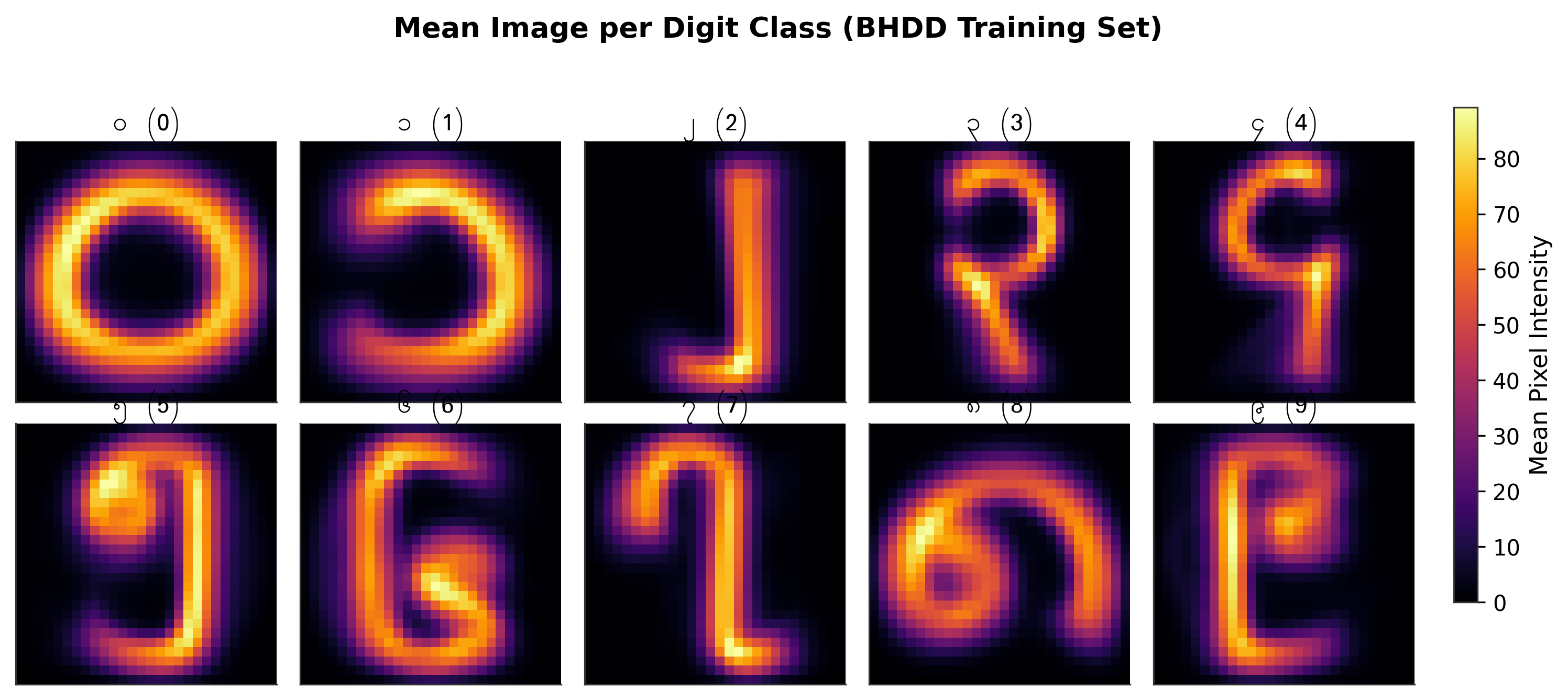}
  \caption{Mean image per class (6,000 training samples each). Sharp
  shapes show consistent digit forms across writers.}
  \label{fig:mean_digits}
\end{figure}

\textbf{Mean images.} Averaging all 6,000 training samples per class
gives the images in Fig.~\ref{fig:mean_digits}. Each class has a
distinct shape: class~0 is a ring, class~1 is an open arc, class~8 is a
spiral. The averages are sharp rather than blurry, meaning writers agree
on the basic form of each digit.

\begin{figure}[!t]
  \centering
  \includegraphics[width=\columnwidth]{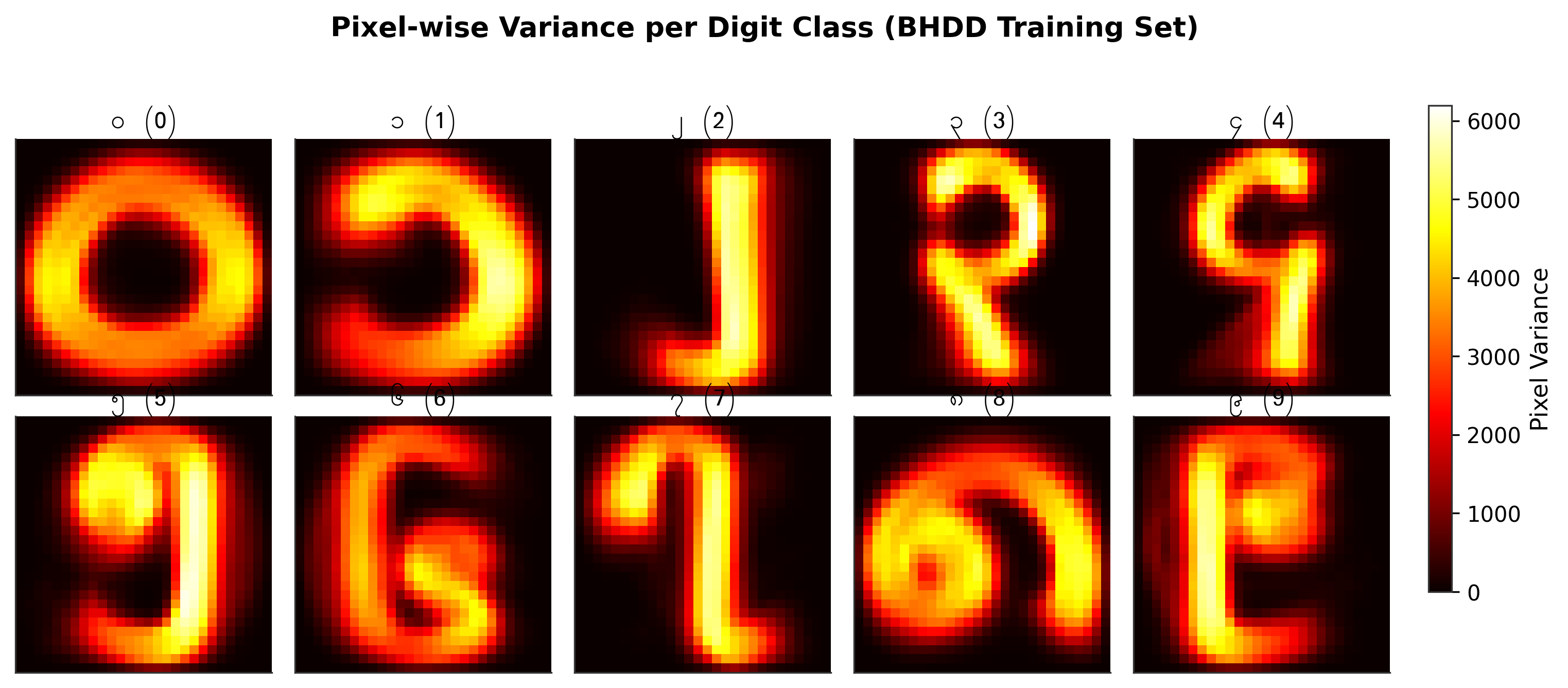}
  \caption{Per-class pixel variance. Bright areas mark where writers
  differ most.}
  \label{fig:variance}
\end{figure}

\textbf{Variance.} Fig.~\ref{fig:variance} maps where pixel values vary
most across writers. High-variance spots sit at stroke endpoints and
junctions, where individual handwriting styles diverge most. Complex
digits like class~8 show wide variance regions; simple ones like
class~2 show a narrow band.

\begin{figure}[!t]
  \centering
  \includegraphics[width=\columnwidth]{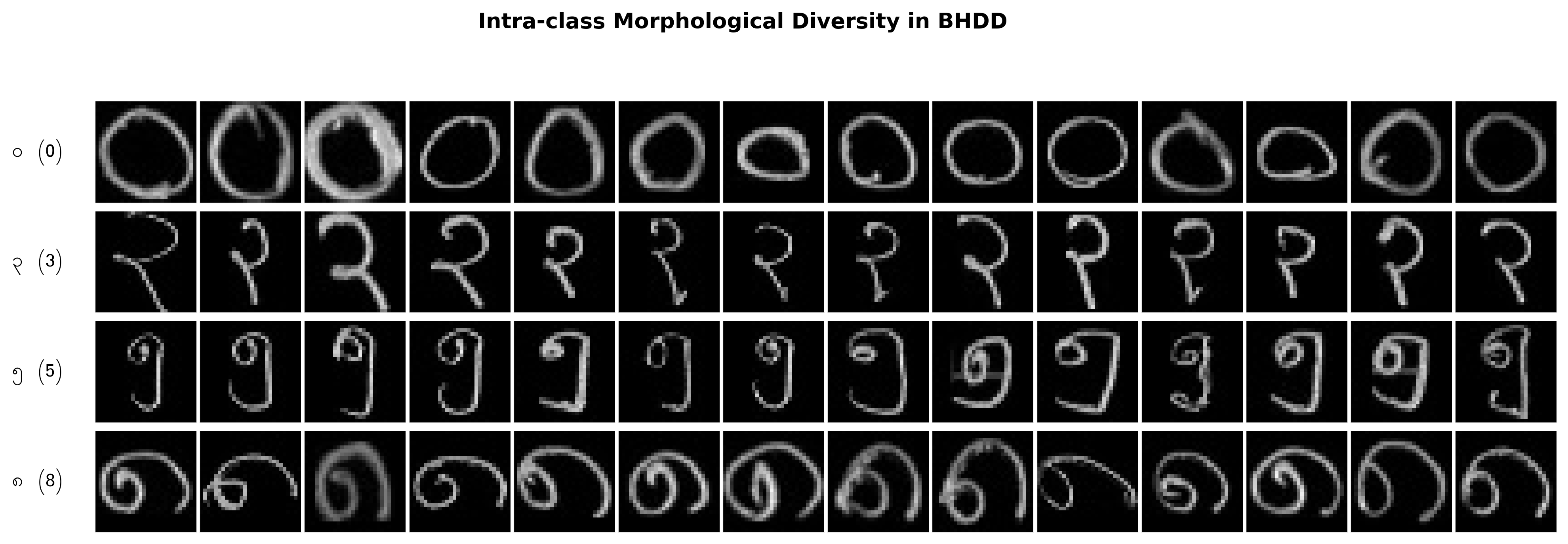}
  \caption{Writing style variation within classes 0, 3, 5, and 8.}
  \label{fig:morphology}
\end{figure}

\textbf{Writing styles.} Fig.~\ref{fig:morphology} shows individual
samples from four classes with high within-class variation (0, 3, 5, 8).
Stroke thickness, curvature, and how far the loop closes all differ from
writer to writer.

% ---- 3.5 ----
\subsection{Script-Specific Challenges}

\begin{figure}[!t]
  \centering
  \includegraphics[width=\columnwidth]{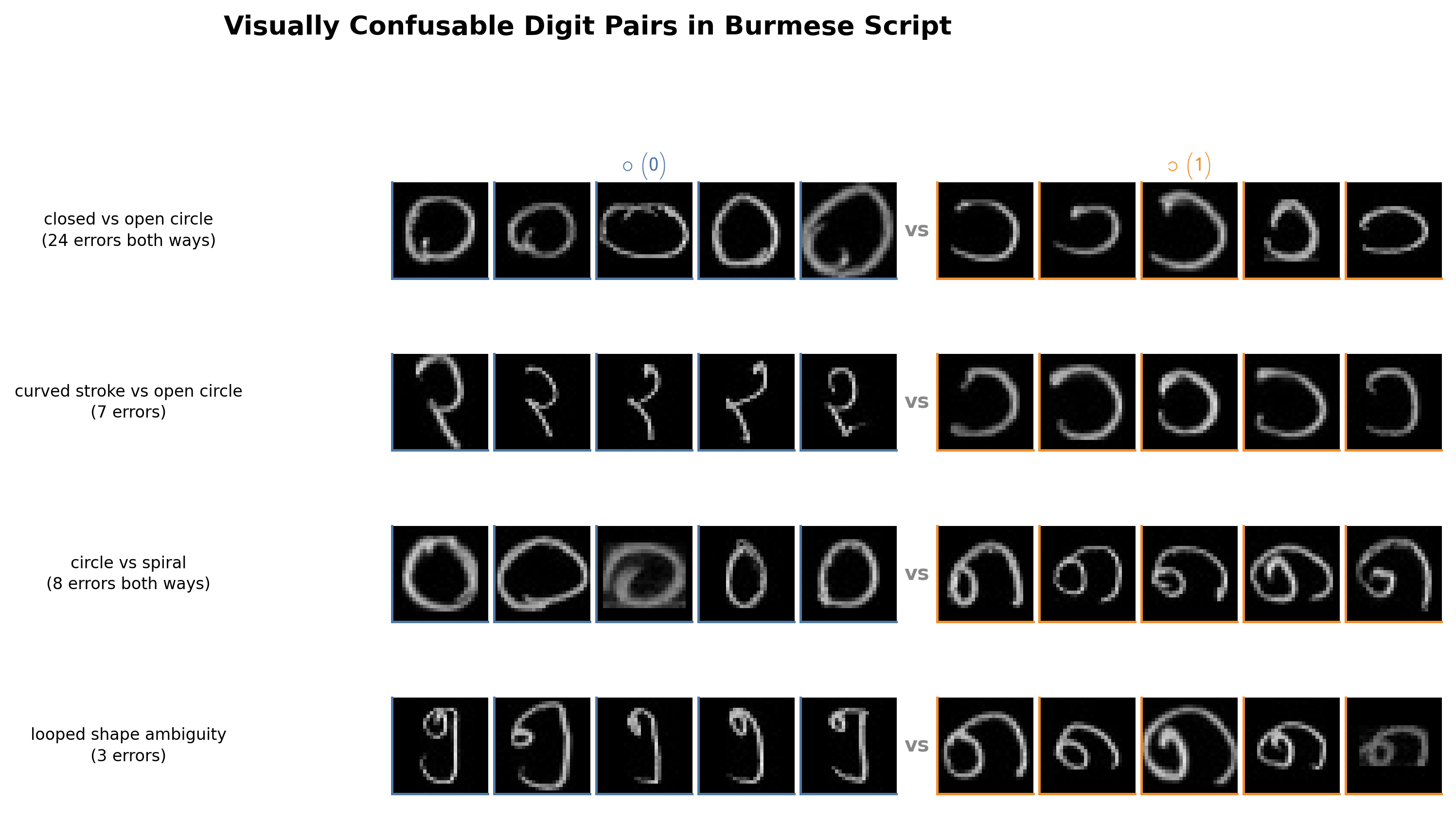}
  \caption{Commonly confused digit pairs with error counts from the improved CNN.}
  \label{fig:similar_pairs}
\end{figure}

Because Myanmar digits are built from round strokes, some pairs look
very similar. We identified the worst pairs from our improved CNN's confusion
matrix (Section~\ref{sec:error}) and show them in
Fig.~\ref{fig:similar_pairs}.

Digits~0 and~1 are the hardest pair: 24 misclassifications between
them (in both directions combined). The only difference is whether the
circle is closed or has a small gap. Digits~0 and~8 share a round outer
shape but differ inside (plain ring vs.\ spiral), with 8 errors
combined. Digit~3 gets mistaken for~1 seven times when its tail is faint
and only the curved head shows. Digits~5 and~8 sometimes overlap because
both have looped forms (3~errors).

When digits share curved sub-strokes, small differences in pen pressure
or whether someone fully closes a loop can shift a sample's appearance
toward a neighboring class.

% ------------------------------------------------------------------
\section{Baseline Experiments}

We trained three models to establish initial performance on the
dataset. All runs used seed~42.

\subsection{Models}

\textbf{MLP.} Two hidden layers (256, 128 units), ReLU, Adam, trained on
flattened 784-d input with early stopping (10\% validation split, up to
50 epochs). Implemented in scikit-learn~\cite{pedregosa2011scikit}.

\textbf{CNN.} Two conv layers (32 and 64 filters, $3\times3$, ReLU,
$2\times2$ max-pool), dropout (0.25 spatial, 0.5 dense), one FC layer
(128 units), 10-way output. Trained 15~epochs with Adam
($\text{lr}=10^{-3}$) and cross-entropy. Implemented in
PyTorch~\cite{paszke2019pytorch}.

\textbf{Improved CNN.} Three conv layers (32, 32, 64 filters) with
batch normalization after each, two max-pool stages, and a 128-unit FC
layer (431K parameters, barely more than the baseline's 421K). Trained
25~epochs with on-the-fly augmentation (rotation $\pm$15\textdegree,
translation $\pm$2\,px, scale 0.9--1.1$\times$), cosine learning-rate
annealing, and light weight decay.

\subsection{Results}

All three models exceed 99\% accuracy with standard training
procedures (Table~\ref{tab:results}). The improved CNN reaches 99.83\%
with a macro F1 of 0.998, cutting misclassifications by about a third
compared to the baseline CNN while adding only 10K parameters.

\begin{table}[!t]
  \centering
  \caption{Baseline Results on the BHDD Test Set}
  \label{tab:results}
  \begin{tabular}{lcccc}
    \toprule
    Model & Accuracy & F1 & Precision & Recall \\
    \midrule
    MLP          & 0.9940 & 0.9905 & 0.9876 & 0.9934 \\
    CNN          & 0.9975 & 0.9964 & 0.9959 & 0.9970 \\
    Improved CNN & 0.9983 & 0.9980 & 0.9972 & 0.9988 \\
    \bottomrule
  \end{tabular}\\[2pt]
  {\small F1, Precision, and Recall are macro-averaged.}
\end{table}

\subsection{Error Analysis}
\label{sec:error}

\begin{figure}[!t]
  \centering
  \includegraphics[width=\columnwidth]{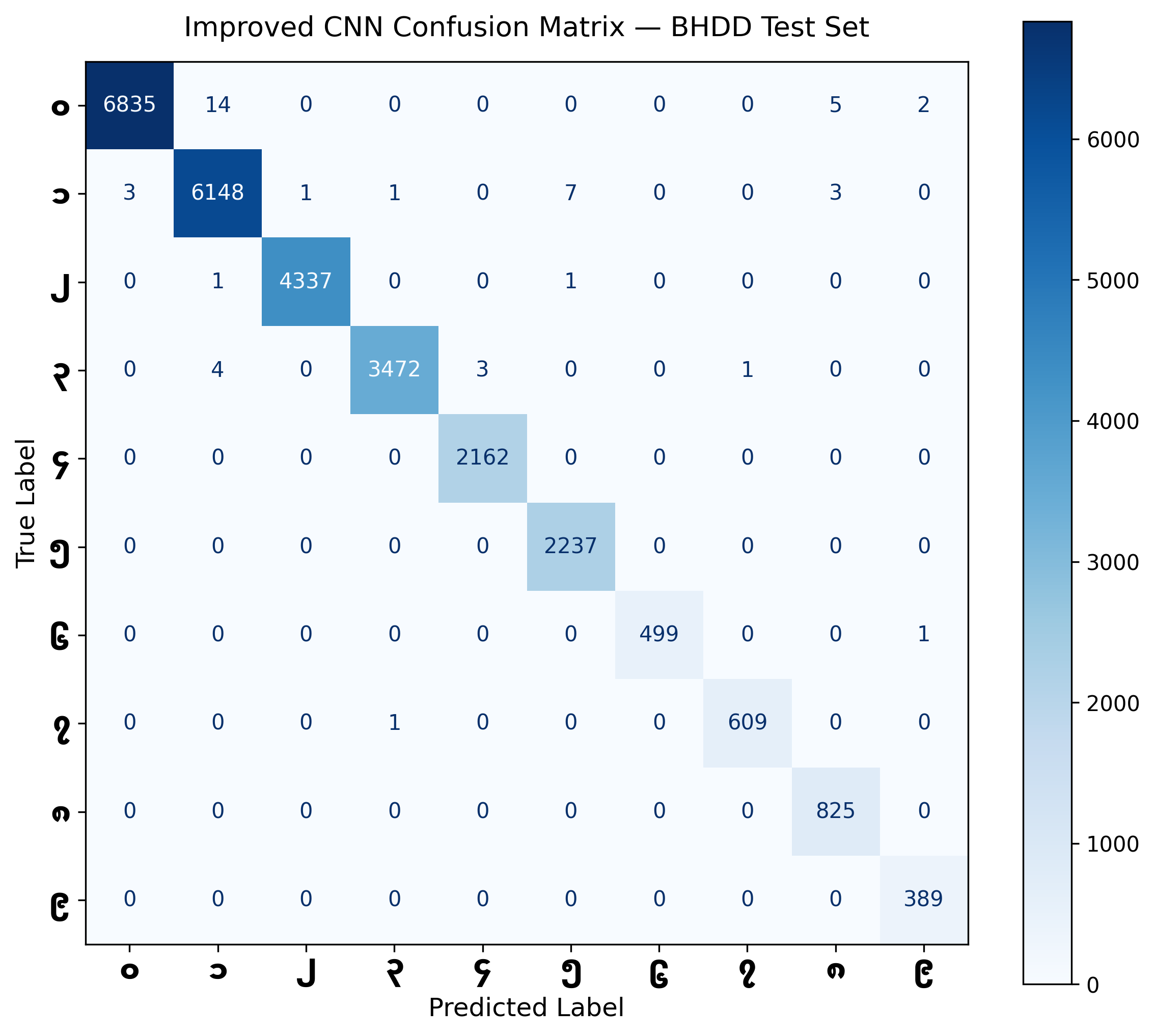}
  \caption{Improved CNN confusion matrix. Nearly all off-diagonal
  entries are zero. The 0--1 pair dominates the remaining errors.}
  \label{fig:confusion}
\end{figure}

The improved CNN's confusion matrix (Fig.~\ref{fig:confusion}) has
nearly all off-diagonal entries at zero. Of 27,561 test samples, only
47 are misclassified. The errors
cluster around digits~0 and~1 (a closed versus open circle), which
account for about half of all mistakes, consistent with the
structural similarity between these two classes.

% ------------------------------------------------------------------
\section{Availability}

BHDD is at \url{https://github.com/baseresearch/BHDD} under CC BY-SA 4.0,
in pickle and IDX formats. The repository includes exploration scripts,
baseline code, and usage examples.

% ------------------------------------------------------------------
\section{Conclusion and Future Work}

We have released BHDD, a dataset of 87,561 Burmese handwritten digit
images from over 150 contributors. The training set is balanced; the
test set preserves the class frequencies as they arose during
collection. Simple baselines reach up to
99.83\% accuracy, and the remaining errors arise from visual similarity between
round digits that share curved sub-strokes.

Next, we plan to expand the dataset to cover Myanmar consonants and
characters, collect word- and sentence-level handwriting, and study how
test-set imbalance affects evaluation.

% ------------------------------------------------------------------
\section*{Acknowledgments}

We thank the volunteers from Taungoo Computer University, the students
of St.\ Augustine / B.E.H.S~(2) Kamayut, and everyone who contributed
handwriting samples.

% ------------------------------------------------------------------
\bibliographystyle{IEEEtran}
\bibliography{references}

\end{document}